\documentclass[review]{elsarticle}

\usepackage{setspace}

\usepackage{lineno,hyperref}
\modulolinenumbers[5]
\usepackage{times}
\usepackage{epsfig}
\usepackage{graphicx}
\usepackage{amsmath}
\usepackage{amssymb}  
\usepackage{stfloats}
\usepackage{bm}
\usepackage{multirow}
\usepackage{latexsym}
\usepackage{threeparttable}
\usepackage{xcolor}
\usepackage{xspace}
\usepackage{footmisc}
\usepackage{makecell}
\usepackage{bbding}
\usepackage{float}
\usepackage{framed}
\usepackage{subfig}
\usepackage{pifont}
\usepackage{adjustbox}

\usepackage{booktabs}
\usepackage{algorithm}
\usepackage{algorithmic}
\usepackage{tablefootnote}
\usepackage[capitalize]{cleveref}
\crefname{section}{Sec.}{Secs.}
\Crefname{section}{Section}{Sections}
\Crefname{table}{Table}{Tables}
\crefname{table}{Tab.}{Tabs.}

\journal{Pattern Recognition}

\newcommand{\ourmodel}{{Generative Compositor}\xspace}
\newcommand{\ourmodelshort}{{GC}\xspace}
\newcommand{\ourmodelseqshort}{{GC\textsubscript{seq}}\xspace}
\newcommand{\encoder}{{Source Encoder}\xspace}
\newcommand{\decoder}{{Target Generator}\xspace}
\newcommand{\resampler}{{Prompt-Aware Resampler}\xspace}
\newcommand{\unimatcher}{{Generative Matcher}\xspace}

\newcommand{\seqmatcher}{{Generative Matcher\textsubscript{seq}}\xspace}
\newcommand{\seqmatchershort}{{GM\textsubscript{seq}}\xspace}

\newcommand{\sod}{{Search One Direction}\xspace}
\newcommand{\sad}{{Search All Directions}\xspace}

\newcommand\mypara[1]{\vspace{1.0mm}\noindent\textbf{#1}}

\newcommand\token[1]{\noindent\texttt{{[#1]}}}

\newcommand\rankfirst[1]{\textbf{#1}}
\newcommand\ranksecond[1]{\underline{#1}}

\let\oldequation\equation
\let\oldendequation\endequation

\definecolor{ForestGreen}{RGB}{34, 139, 34}

\renewenvironment{equation}
{\linenomathNonumbers\oldequation}
{\oldendequation\endlinenomath}

\begin{document}
\begin{sloppypar}

\begin{frontmatter}

\title{Generative Compositor for Few-Shot Visual Information Extraction}

\author{Zhibo Yang$^{a,b\ast}$}
\ead{yangzhibo450@gmail.com}

\author{Wei Hua$^{a\ast}$}
\ead{Perdredes@outlook.com}

\author{Sibo Song$^{b\ast}$}
\ead{sibosongzju@gmail.com}

\author{Cong Yao$^{b}$}
\ead{yaocong2010@gmail.com}

\author{Yingying Zhu$^{a}$}
\ead{yyzhu@hust.edu.cn}

\author{Wenqing Cheng$^{a}$\textsuperscript{\ding{41}}}
\ead{chengwq@hust.edu.cn}

\author{Xiang Bai$^{a}$}
\ead{xbai@hust.edu.cn}

\address{a. Huazhong University of Science and Technology, Wuhan, Hubei, China}

\address{b. Alibaba Group, Hangzhou, Zhejiang, China}
\cortext[cor]{Equal contribution. \ding{41}~Corresponding author}

\begin{abstract}
Visual Information Extraction (VIE), aiming at extracting structured information from visually rich document images, plays a pivotal role in document processing. Considering various layouts, semantic scopes, and languages, VIE encompasses an extensive range of types, potentially numbering in the thousands. However, many of these types suffer from a lack of training data, which poses significant challenges. 
In this paper, we propose a novel generative model, named Generative Compositor, to address the challenge of few-shot VIE. 
The Generative Compositor is a hybrid pointer-generator network that emulates the operations of a compositor by retrieving words from the source text and assembling them based on the provided prompts. 
Furthermore, three pre-training strategies are employed to enhance the model's perception of spatial context information. Besides, a prompt-aware resampler is specially designed to enable efficient matching by leveraging the entity-semantic prior contained in prompts. The introduction of the prompt-based retrieval mechanism and the pre-training strategies enable the model to acquire more effective spatial and semantic clues with limited training samples. Experiments demonstrate that the proposed method achieves highly competitive results in the full-sample training, while notably outperforms the baseline in the 1-shot, 5-shot, and 10-shot settings. 
\end{abstract}


\begin{keyword}
Visual Information Extraction \sep Few-Shot \sep OCR
\end{keyword}

\end{frontmatter}


\section{Introduction}
\label{sec:intro}

The Visual Information Extraction (VIE) task aims to extract structured information from document images~\cite{toledo2019information,wang2021towards}, particularly those with complex structures and elements, such as customized receipts, tabular bills, and nested tables. VIE possesses a significant research value when handling documents with intricate visual representations and semantics, which implies that multi-modal approaches may serve as effective solutions. In the study of VIE, traditional methods to handling multi-modal information complement morphological tagging with keywords indexing, while recent models explore graphs ~\cite{liu2019graph}, multi-modal fusion~\cite{xu2020layoutlm}, and knowledge embedding~\cite{peng2022ernie}.

\begin{figure}[!t]
\begin{minipage}{0.52\linewidth}

    \centering
    {\includegraphics[width=0.97\linewidth]{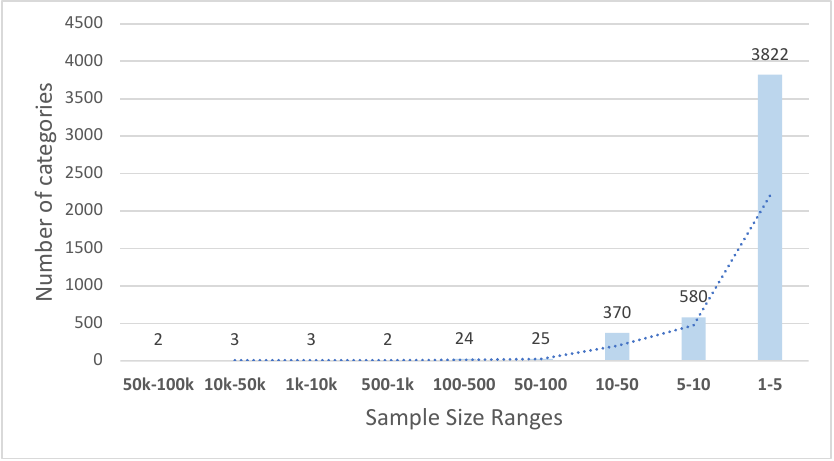}%
    }
    
    \caption{\textbf{Number of categories under different sample size ranges}. The statistics\protect\footnotemark{} shows that there are 3,822 categories with sample sizes less than 5, while no more than 3 categories with ranges higher than 500.}
    \label{fig:statistics}
\end{minipage}
\hfill
\begin{minipage}{0.46\linewidth}

    \centering
   {\includegraphics[width=1\linewidth]{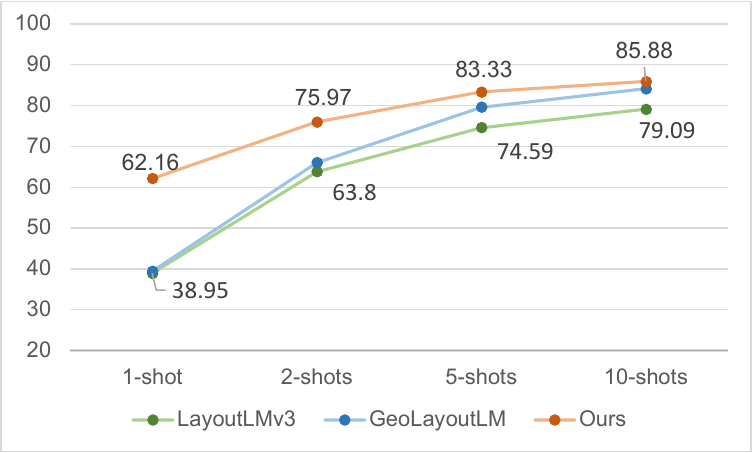}%
    }
    
    \caption{\textbf{Few-shot comparisons on CORD}. Our method outperforms LayoutLMv3 and GeoLayoutLM in all four few-shot settings, with a significant advantage in the 1-shot setting.}    
    \label{fig:few-shot-comp}
\end{minipage}
\end{figure}
\begin{figure}[!t]
\centering
\captionsetup[subfloat]{font=scriptsize}
{\includegraphics[width=1\linewidth]{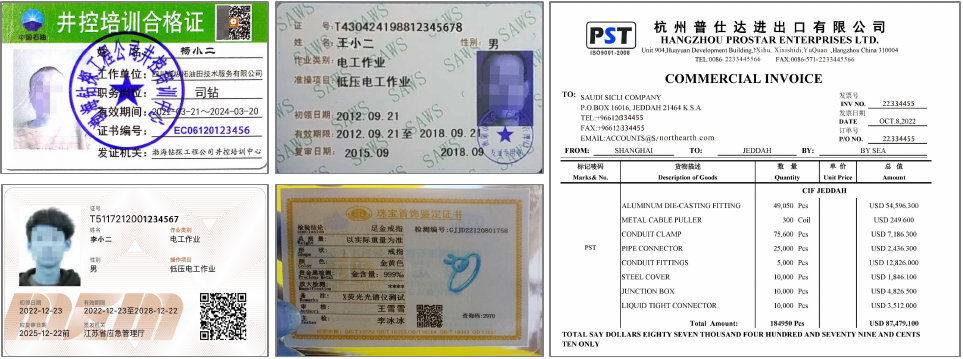}}%
\caption{\textbf{Show case of categories with sample size less than 5 in~\cref{fig:statistics}}.}
\label{fig:intro_few_shot}

\end{figure}

In practical applications, VIE often confronts scenarios with limited samples, yet necessitates a high level of accuracy. On one hand, VIE often deals with objects that have rich visual layouts and semantic scopes. As a result, most scenarios have only a small amount of training data available. As shown in~\cref{fig:statistics}, we present a severe imbalance in the VIE scenario, with a significant long tail phenomenon (showcases are presented in~\cref{fig:intro_few_shot}). On the other hand, the information to be extracted is often utilized for purposes such as verification, auditing, and archiving, allowing for small accuracy errors. A typical VIE algorithm comprises several interconnected components, including OCR, structural analysis, and entity labeling and linking. In such a pipeline, two challenges impact the accuracy: 1) The scarcity of samples makes it difficult to learn multi-modal features: textual, visual, and layout. 2) The errors introduced by the offline OCR engines might be a bottleneck for downstream entity labeling and linking. LayoutLM~\cite{xu2020layoutlm} and its variants~\cite{xu2021layoutlmv2,huang2022layoutlmv3} adopt offline OCR engines and formulate VIE as a token classification problem. To better cope with the discrepancy between visual and spatial features, they resort to \textit{document layouts}, a kind of geometric clue of texts, to assist models to automatically construct relations. However, layouts are sensitive to the text reading order~\cite{qiao2024reading}, which is often incorrect in skewed or multi-column documents. TRIE~\cite{zhang2020trie} and ESP~\cite{yang2023modeling} are end-to-end trainable frameworks that integrate OCR and IE. These end-to-end methods often require the design of specialized links at the decoding stage to address the complex structural relations between entities. The latest approaches utilize generative methods to directly generate a sequence of discrete tokens that represent the entities and their relations. UDOP~\cite{tang2023unifying} and Dount~\cite{kim2022donut} are representative methods of generative models that take an image as input and directly output the sequence containing text recognition and entity extraction results. However, such methods would occasionally generate unstable or hallucinated results, which can not be directly diagnosed because of the black-box nature of their methodology.

\footnotetext{https://duguang.aliyun.com/experience?type=docAutoStudy. The data utilized corresponds to the VIE machine learning platform from Alibaba Cloud in the fourth quarter of 2023.}
To tackle these challenges, we propose a generative method called \textit{\textbf{Generative Compositor}} (\textbf{GC} for short), as illustrated in~\cref{fig:main_architecture}. The core idea of our method can be summarized by the terms ``generative" and ``compositor", motivated by the Pointer Generator Network~\cite{see2017get} proposed for the Summarization task in NLP. The term ``generative" refers to the process of generating vectors that point to OCR text blocks, where the values of these vectors are the most likely B-I-O categories under the queries ( namely, keys in VIE ). The term ``compositor" involves rearranging the OCR text blocks by their B-I-O categories to produce the final complete VIE values. Taking advantage of multi-modal pre-training~\cite{xu2020layoutlm}, GC starts with a source pool, in which each element is represented by informative features from three modalities: vision, geometry, and OCR. The target of GC is a structured sequence that answers the corresponding question, i.e., \textit{``the value of discount is?''}. Unlike GenKIE~\cite{cao2023genkie}, our approach relies on the OCR recognition results and will not tamper the content recognized by the OCR engine. Based on our observations, OCR generally achieves satisfactory recognition accuracy for documents, particularly in scanned documents. In real applications, another issue that actually affects the VIE performance more seriously stems from the arrangements of text blocks, which are frequently disturbed by various factors: table cells, separated paragraphs, and printing quality (e.g., rotation and shift). Considering these challenges, our method does not generate the final results in an end-to-end manner. Instead, it generates category vectors, and then composites the disordered OCR fragments based on these vectors to form final answers. This approach not only trusts the OCR results but also addresses the issue of disorder. Therefore, this method is more applicable for VIE, which requires precise control over the formats and contents of the outputs, rather than general Visual Question Answering (VQA), which encourages rich, natural language-based answers.  


To enhance the model's perception of spatial context information, we further propose three pre-training strategies. As discussed in ERNIE-Layout~\cite{peng2022ernie},  synthetic document generator~\cite{raman2022synthetic}, and Hi-VT5~\cite{tito2023hierarchical} layout information and spatial clues play a crucial role in visual document understanding. Inspired by these works, we devise pre-training tasks to learn reading orders and linking relations between keys and values: 1) \textit{Match to Fill (MF)}: Inspired by MLM~\cite{xu2020layoutlm}, MF predicts the content between any two words in sequential order. MF encourages the model to pay attention to reading order. 2) \textit{Search for One Direction (SOD)} searches the k nearest words in any one of four directions (Up, Down, Left, Right) from a given entity. SOD enhances the model's perception of directional linking between entities. 3) \textit{Search for All Directions (SAD)} extends the 1-D directional prior to 2-D, aiming to enable the model to learn directional priors that span from lines to region. SAD enables the model to better adapt to information extraction in multiple directions. 

The proposed method addresses the few-shot VIE problem in two ways. Firstly, it transforms entity classification into a sequential matching problem, leveraging contextual information. Secondly, the model employs LayoutLM encoder, and the pre-training tasks devised in this paper to learn better multi-modal features.
Experiments demonstrate that Generative Compositor not only achieves state-of-the-art (SOTA) performance on public VIE datasets but also exhibits excellent few-shot capability. Notably, our method significantly outperforms previous methods by a large margin in 1-shot, 5-shot, and 10-shot experiments, as shown in ~\cref{fig:few-shot-comp}. The contributions of this work are as follows:
  
\begin{itemize}
    \item We propose a novel method for VIE, called Generative Compositor, which leverages the prior knowledge of layout and prompts, enabling more effective few-shot learning.
    \item We devise three pre-training tasks to improve the model's perceptual capacity for spatial context information.
    \item We introduce a prompt-aware resampler to enable efficient matching in the decoding stage by distilling the multi-modal embedding. 
    \item Extensive experiments demonstrate that the proposed method achieves competitive results in the full-sample setting and a significant advantage in few-shot settings.
\end{itemize}


\section{Motivation}
\label{sec:motivation}
Learning informative representation from limited samples in the encoding phase, and employing a suitable decoding strategy in the decoding phase, are important for few-shot visual information extraction. 

Within the representation learning phase, our objective is to comprehensively capture contextual information from diverse modalities. Contextual information in VIE can be categorized into two types: visual context, which encompasses the layout of the document, and linguistic context, which involves the interconnections between the target content and other textual elements. Previous works such as LayoutLMv3~\cite{huang2022layoutlmv3}, UniDOC~\cite{gu2021unidoc}, and GeoLayoutLM~\cite{luo2023geolayoutlm} have demonstrated the following inequality across various encoders: texts only \textless texts + layout \textless texts + layout + image embeddings. As shown in~\cref{tab:main_full_shot}, when using only text tokens, the best performance is achieved by BERT\textsubscript{large}, with an F1 score of 65.6 on the FUNSD dataset~\cite{jaume2019funsd}. Incorporating layout information improves performance significantly, with BROS and LiLT~\cite{wang2022lilt} achieving an F1 score higher than 80 on FUNSD. However, methods leveraging multi-modal inputs (texts, layout, and image embeddings) further boost the F1 score higher than 90.

For visual context, the proposed method leverages pre-trained layout embedding, inspired by the advancements made in the LayoutLM~\cite{huang2022layoutlmv3}. However, it diverges from LayoutLM in the treatment of linguistic context. In addressing linguistic context, we employ a simple yet effective prompting strategy: $ < $\textit{What is the name of the menu}$ ?>, < $\textit{What is the count of
the menu}$ ?>, < $\textit{...}$ ?> $. This straightforward strategy constructs the simplest key-value context, where the model can link keys to values with specific attributes (such as time, quantity, and price) with only a single sample.

In the design of the decoder, we adopt a prompt-matching mechanism instead of a conventional classification-based approach. Classification-based methods must specify the number of categories, and when there are very few samples, unseen categories cannot be learned. In contrast, a matching-based approach can still match the most probable instance for unseen categories. Besides, the prompts not only give the targets for matching but also enrich the linguistic representations, as previously discussed.

\section{Related Work}
\label{sec:related}
The recent advent of deep-learning-based approaches has led to remarkable progress in using layout-induced representations in visual document understanding. 
A multitude of methods have been proposed recently to tackle the visual information extraction (VIE) challenge, which is one of the most prominent applications of visual document understanding.
Existing VIE methods can be roughly separated into three categories: classification-based Models, linking-based Models, and generation-based models.

\mypara{Classification-based Models.}
Most of the previous works use a classification approach for entity recognition, with the difference being whether OCR is treated as an external module or not. On one hand, the methods adopt OCR as an external input focus on building layout-aware or graph-based representation for documents and seeking to achieve VIE via classification-based methods. 
The representative works adopt the sequence labeling paradigm to classify text tokens, including LayoutLM family~\cite{xu2020layoutlm}, DocFormer~\cite{appalaraju2021docformer}, SelfDoc~\cite{li2021selfdoc}, UniDoc~\cite{gu2021unidoc}, ERNIE-Layout~\cite{peng2022ernie}, GeoLayoutLM~\cite{luo2023geolayoutlm}. The innovation of these works lies in novel pre-training tasks or new cross-modal transformer layers. These pre-training works with the goals of spatial awareness and image-text alignment have greatly advanced the task of visual information extraction, and their ideas have also inspired the research on few-shot learning in this paper. 
On the other hand, some models use image-only inputs~\cite{zhang2020trie} by integrating OCR into the whole framework. These models allow for synergistic optimization from text spotting to information extraction. 
However, these methods rely on text with proper reading order or an extra module~\cite{zhang2023reading} for serialization. StrucTexTv2~\cite{yu2022structextv2} and ESP~\cite{yang2023modeling} leverage segment-level classification for text blocks, resulting in a lack of ability for token-level extraction. 
Although the joint learning methods can address OCR errors and enable feature sharing between multi-modal encoder and OCR encoder, an internal OCR is not necessary in most scanned documents where the precision of OCR is high enough.

\mypara{Linking-based Models.}
Linking and grouping aim to extract the relations between entities. Linking primarily focuses on key-value pairs, while grouping is used for combining multiple values into a single item. There has been an increasing interest in studying linking-based techniques. BROS~\cite{hong2022bros} conducts a binary classification for all pairs of candidate tokens, resulting in high computational costs. 
To address the aforementioned reading order issue, SPADE~\cite{hwang2021spatial}, and ESP~\cite{yang2023modeling} propose serializer-free models by introducing extra links for modeling complex relations, for example, intra-group or inter-group relations.
TPP~\cite{zhang2023reading} model proposes a Token Path Prediction module to model all document tokens as a complete directed graph of tokens. GraphRevisedIE~\cite{cao2023graphrevisedie} links entities by a revised graph.
Although these methods employ extra links or modules to solve the reading order issue, the complicated decoding or post-processing strategy limits their generalizability. In the proposed work, we establish the KV relations by answering prompts, i.e., "What is the name of the menu?" and "What is the count of the menu?". In this way, we convert linking to questioning and answering. 

\mypara{Generation-based Models.}
These methods can be roughly divided into two categories: OCR-free methods and OCR-dependent methods.
Generally, generation-based models are capable of performing a variety of extraction tasks by simply homogenizing outputs into a structural sequence of discrete tokens.
Recent end-to-end models remove the reliance on OCR tools and adopt schema-based outputs to avoid sophisticated processing. Donut~\cite{kim2022donut} is the first work that directly maps an input document image into a desired structured output. Dessurt~\cite{davis2022end} is another generative method that is capable of performing text recognition and document understanding. The structured generation module in ~\cite{cao2022query} is able to generate the final key-value pairs even with OCR noise.  
Other methods~\cite{tang2023unifying,cao2022query} utilize external OCR engines and generate entity types or content given prompts or learnable queries, without requiring any serialization steps. However, they lack localization ability and faithfulness to the original image text.

The method we propose can be categorized as OCR-dependent generative methods. By matching, fetching, and assembling words from source OCR text, GC generates stable, interpretable, and structured sequences with minimal post-processing. Classification-based and linking-based methods always require additional post-processing to form final structured outputs. Generation-based models lack interpretability, as no grounding information is given. They may also produce unstable or hallucinated results. Similar to the pointer mechanism in PGN~\cite{see2017get}, our model copies the most probable answer from the OCR text, and meanwhile inserts delimiters to create a structured representation of the results.


\section{Methodology}
\label{sec:method}
In this section, we first present the overall paradigm of our method. Unlike previous generation-based methods~\cite{kim2022donut, cao2023genkie} that directly generate extraction results, our approach learns to generate vectors for matching the correct output tokens. 
In the following subsections, we detail the components of our model, including the Source Encoder, Target Generator, as well as
our core designs, the Prompt-Aware Resampler and the Generative Matcher.
Finally, we describe three pre-training strategies that are specifically designed to learn spatial-aware and semantic-aware representations.

\begin{figure*}
    \vspace{-2mm}
    \centering 
    \centerline{\includegraphics[width=1.\linewidth]{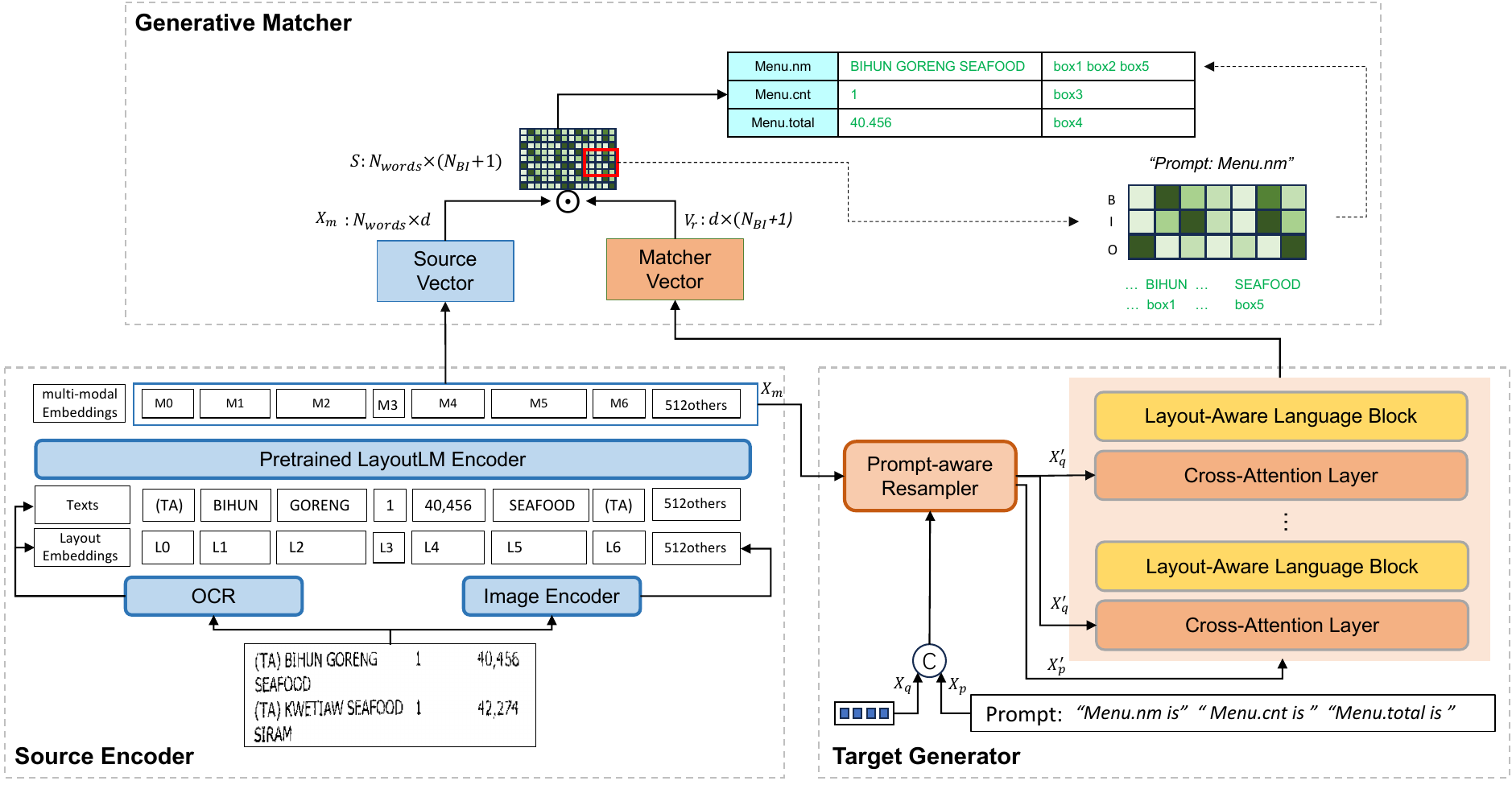}}
    \caption{\textbf{A schematic illustration of the proposed Generative Compositor}. Given a document image and a prompt, the Generative Matcher produces answers with grounding information, by calculating the similarity between the source vector encoded by the Source Encoder and the matcher vector generated by the Target Generator.}
    \label{fig:main_architecture}
\end{figure*}

\subsection{Model Architecture}
As illustrated in ~\cref{fig:main_architecture}, our \ourmodel can be summarized into three parts: Source Encoder, Target Generator, and Generative Matcher.
The Source Encoder takes a document image as input and produces source multi-modal embeddings $X_m \in \mathbb{R}^{N_{words} \times d}$ processed by OCR, an image encoder and a pre-trained LayoutLM encoder, where $N_{words}$ denotes the number of source text tokens.
Note that the pre-trained LayoutLM encoder and Layout-Aware Language Block in the generator both take multi-modal inputs containing visual, textual, and layout embeddings, which can be initialized with many existing Transformer-based document pre-trained models~\cite{huang2022layoutlmv3}.
Next, the Prompt-Aware Resampler receives multi-modal embeddings $X_m$ from the Source Encoder and concatenated queries of prompt embeddings $X_p \in \mathbb{R}^{N_p \times d}$ and learnable queries $X_q \in \mathbb{R}^{N_q \times d}$, where $N_p$ and $N_q$ denote the number of prompt tokens and learnable queries respectively.
After that, as shown in ~\cref{fig:resamplar}, the Prompt-Aware Resampler outputs enhanced resampled embeddings ${X_q}'$ and enhanced prompt embeddings ${X_p}'$.
Later, the Target Generator is fed with the resampler outputs and produces vectors $V_r \in \mathbb{R}^{d \times (N_{BI}+1)}$ to match the multi-modal embeddings $X_m$, where $N_{BI}+1$ denotes total number of entity tags.
Finally, the Generative Matcher calculates a similarity matrix $S$ by multiplying $X_m$ by $V_r$, which is then used to achieve entity classification.
Note that an alternative \seqmatcher is devised to better solve the issues of OCR orders in practical applications (see~\cref{sec:matcher}). 

\subsubsection{Source Encoder and Target Generator}
\label{sec:encoder}
To fully exploit the visual and layout information, we employ the widely recognized LayoutLMv3~\cite{huang2022layoutlmv3} pre-trained backbone to initialize the encoder and decoder modules.
During training and inference, the document image is first resized and split into a sequence of uniform patches.
Afterward, the document images are represented with patch features via linear projection layers.
Moreover, the text and layout information can be extracted via an off-the-shelf OCR engine and then concatenated with visual patch features before feeding them into the \encoder.

\subsubsection{\resampler}
\label{sec:resampler}

\begin{figure}
    \centering 
    \centerline{\includegraphics[width=.6\linewidth]{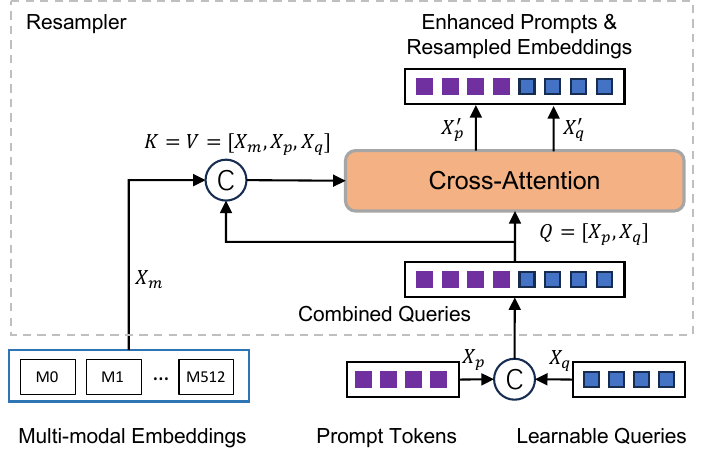}}
    \vspace{-2mm}
    \caption{\textbf{A diagram of \resampler.} }
    \label{fig:resamplar}
\end{figure}

Considering the prevalence of long documents, we employ a resampler-based module, which draws inspiration from Flamingo~\cite{alayrac2022flamingo}, facilitating the adaptive distillation of tokens. In the original Flamingo, the resampler serves only one purpose: to reduce the number of visual tokens. In our study, we aim to achieve two objectives: 1) the reduction of visual encoders should take prompt information into account and 2) the encoding of the prompt itself can be further enhanced.
The vanilla resampler adopts an attentive pooling mechanism for capturing semantics from textural features without attending to the prompt tokens, which makes the distilled features less informative to the prompt tokens.
To remedy this drawback, we propose the \resampler (PAR) (shown in~\cref{fig:resamplar}) to enable a mutual enhancement between prompt tokens and textural features.
Specifically, the \resampler first concatenates the prompt tokens with learnable queries into a combined query. 
With each combined query, the resampler cross-attends the concatenation of multi-modal embeddings $X_m$ and the combined query, as suggested in~\cite{alayrac2022flamingo}.
Formally, the enhanced resampled embeddings ${X_q}'$ and enhanced prompt embeddings ${X_p}'$ are computed as,
\begin{equation}
    \langle {X_q}',{X_p}' \rangle = \mathbf{PAR}\left(\langle {X_q},{X_p} \rangle; X_m \right)
    \label{eq:resampler}
\end{equation}
Flamingo demonstrates that concatenating the features of learnable queries with multi-modal features yields better performance compared to using multi-modal features alone. We attribute this improvement primarily to the increased capacity of feature representation. Consequently, when we incorporate prompt embeddings, we follow the same approach by concatenating them with multi-modal features.
Note that PAR not only provides clear entity-related clues for distilling information from the encoder's output but also enriches the representation of prompt embeddings. 



\subsubsection{Generative Matcher}
\label{sec:matcher}

This section presents the core component of our proposed paradigm, \unimatcher.
To accomplish context-aware token classification for information extraction, a decoder is employed as the generator to produce weights for sequence classification. 
By utilizing these conditionally parameterized weights, our method produces classification weights conditioned on each input, thus differentiating it from the conventional sequence labeling. This mechanism yields exceptional improvement.

\begin{figure}
    \centering \centerline{\includegraphics[width=.8\linewidth]{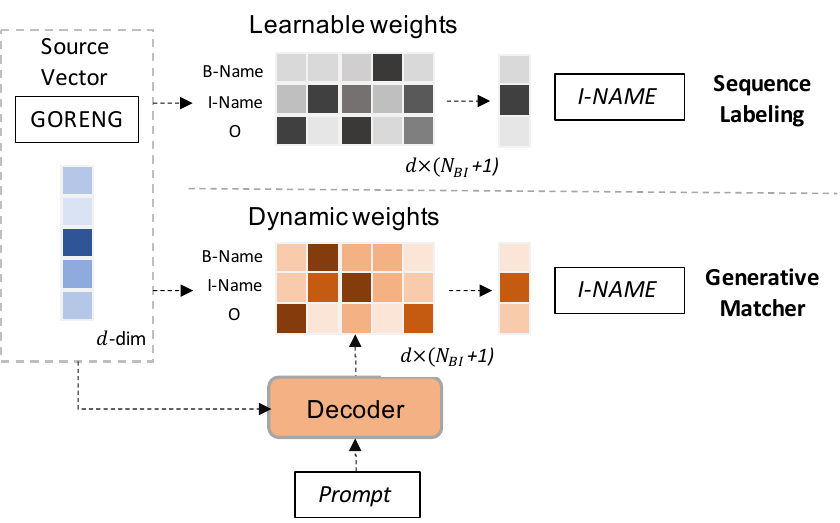}}
    \caption{\textbf{Comparison of Sequence Labeling and our proposed \unimatcher.} Inspired by the dynamic weight design in computer vision methods~\cite{cheng2021per}, we propose to apply the generator to produce the conditioned dynamic weights which serve as the weights of the token classification layer. }
    \label{fig:dynamic}
\end{figure}

\cref{fig:main_architecture} illustrates the position of the Generative Matcher within the overall framework, while \cref{fig:dynamic} details the match vector.
Specifically, the \decoder produces one matcher vector conditioned on each prompt and the resampler output. 

The match-vector is a two-dimensional vector with the size of $(N_{BI}+1) \times d$, where $(N_{BI}+1)$ represents the categories of BIO tagging, and d represents the dimensions of the multi-modal embeddings. The categories of $(N_{BI}+1)$ correspond to the BIO labels of the entity sequence, which include two levels: 1) whether it is a targeted entity, and 2) whether it is the $[beginning]$ or $[inside]$ of the target entity. Therefore, the total number of BIO tagging can be formulated as, $N_{total} = N_{BI} + 1 = N_{B}+N_{I}+1$, where $N_{B}$ represents the number of beginning label of each entity, $N_{I}$ represents the number of inside label of each entity, and 1 represents the outside (non-entity). Typically, $N_{B}$ and $N_{I}$ equals the categories of entities. The total number of BIO tagging in a CORD dataset can be exemplified as, $B-Menu.nm, I-Menu.nm, B-Menu.cnt, I-Menu.cnt, ..., B-Menu.total, I-Menu.total, O$.

Formally, the matching similarity matrix is formulated as $M = X_m \cdot V_r$ where $M \in \mathbb{R}^{N_{words} \times (N_{BI}+1)}$.
Then, the cross entropy (CE) loss for binary classification is introduced as the objective,
\begin{equation}
L = - \sum \left[ Y \cdot \log \sigma(M) + (1 - Y) \cdot \log (1 - \sigma(M)) \right],
\label{eq:uniloss}
\end{equation}
where $Y \in \{0,1\}$ denotes the binary labels, and $\sigma(\cdot)$ denotes the sigmoid function, 

Although the approach bears a resemblance to the sequence labeling technique employing the BIO tagging scheme, \textit{our proposed matching mechanism fundamentally diverges from the conventional sequence labeling.}
As depicted in~\cref{fig:dynamic}, the weights are the same for all input examples in a regular classification layer derived from sequence labeling. 
In contrast, within the \unimatcher, the classification weights are computed through a function that takes into account both the input image and prompts, resulting in higher capability for the classification layer.
\begin{figure}[!t]
\centering
\captionsetup[subfloat]{font=scriptsize}
{\includegraphics[width=0.8\linewidth]{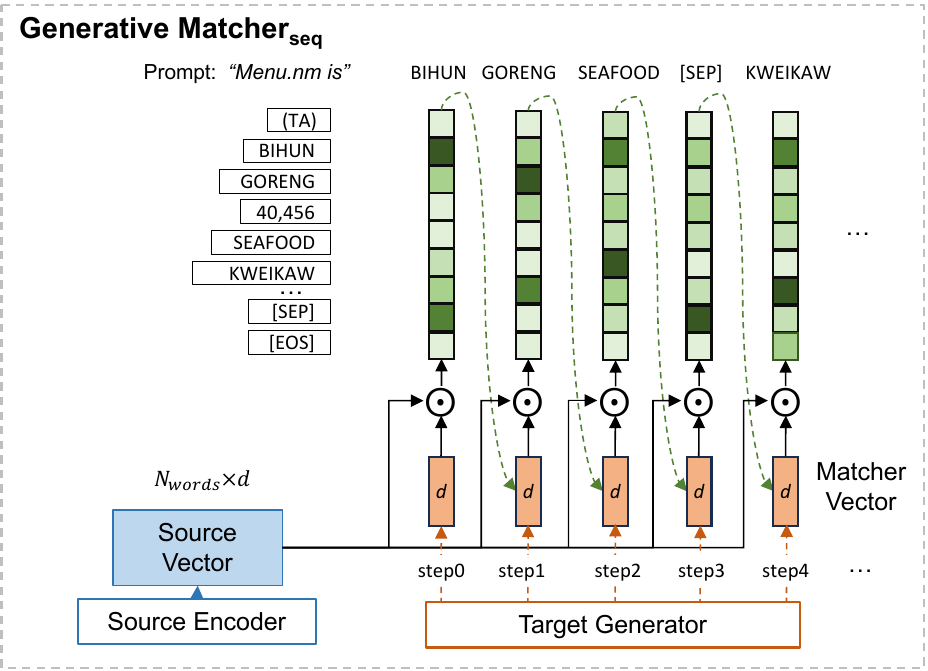}}%
\caption{\textbf{Generative Matcher\textsubscript{seq}:} A sequential variant of Generative Matcher to address the issue of text serialization.}
\label{fig:seq_matcher}
\end{figure}

\mypara{The sequential variant: Generative Matcher\textsubscript{seq} (\seqmatchershort).}
In many real-world scenarios, semantic information is more reliable than spatial positioning. For example, many documents have similar keys but different layouts. Besides, the order of text itself is often misplaced due to the complex layout.
Therefore, we propose a sequential variant of our \unimatcher (see~\cref{fig:seq_matcher}) to produce sequential matcher vectors $V_{seq}$ that match a whole entity in an autoregressive manner. 
If the document contains repeated entities, such as the CORD~\cite{park2019cord} dataset, the \seqmatchershort is trained to match and generate a specialized token of \token{SEP} for separating repeated entities.
\seqmatchershort stops processing when it predicts an end-of-sequence token \token{EOS}.
The two special tokens \token{SEP} and \token{EOS} are added to the input text tokens to facilitate the decoding.
The objective of \seqmatchershort is a standard cross-entropy loss,
\begin{equation}
L=-\sum_{i}^N Y_i \cdot \log \mathtt{Softmax} \left( X_m \cdot V^{i}_{seq} \right)
\label{eq:seqloss}
\end{equation}
where $Y_i$ is the one-hot vector that denotes the source token to match and $V^{i}_{seq}$ denotes the matcher vector at $i^{\text{th}}$ step.
$N$ is the length of the sequence.

\subsection{Pre-training Strategies} \label{subsec_pretrain_intro}

\begin{figure}
    \vspace{-2mm}
    \centering \centerline{\includegraphics[width=.5\linewidth]{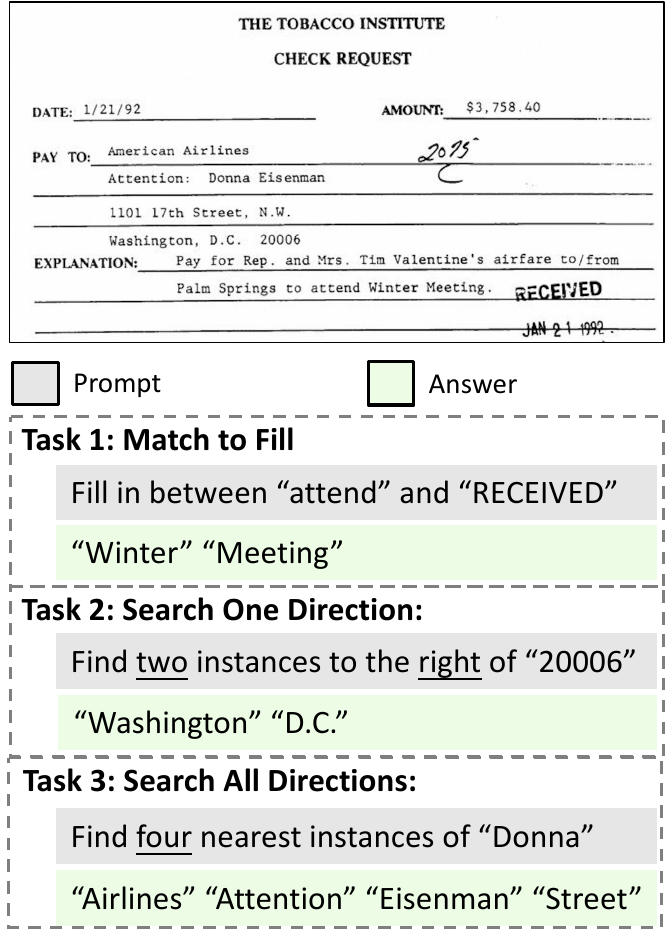}}
    \vspace{-3mm}
    \caption{\textbf{Illustration of pre-training tasks:} Match to Fill, \sod, and \sad, that aim at enhancing spatial relations and learning reading orders. Prompting words with \underline{underline} can be replaced with other settings.}
    \label{fig:pretrain}
    \vspace{-5mm}
\end{figure}

As many document pre-training architectures are not specifically designed for the matching mechanism, we propose three pre-training strategies (see~\cref{fig:pretrain}) for training our proposed paradigm.

\mypara{Match to Fill (MTF).} To encourage the model to match content from contextual text and, more importantly, to learn reading order, we adapt the MTF task to enforce the model to match tokens for filling in blanks between the two given tokens within a single sentence.

\mypara{\sod (SOD).} For VIE tasks, key-value pairs are usually aligned either horizontally or vertically.
Consequently, the acquisition of spatial relations plays a crucial role in effectively extracting key information from the document, which has shown strong potential~\cite{luo2023geolayoutlm} in document understanding.
To address this issue, we propose \sod to facilitate spatial-aware learning via matching the K-nearest words in a specific direction (Left/Right/Top/Bottom) relative to the given word.

\mypara{\sad (SAD).} 
Drawing upon topological heuristics, we propose a \sad (SAD) task that aims to match K-nearest words in all directions. This approach allows the model to enhance its utilization of the spatial layout of keys and values present in the document. 
It is worth noting that SAD complements the SOD, since it solely focuses on learning relations along the x- or y-axis individually, while SAD is applied to enforce the matcher to capture pairwise distances in 2-D space.

\section{Experimental Results}
In this section, we first present the implementation details during pre-training and fine-tuning. 
Subsequently, various datasets employed for training and evaluation are introduced. Finally, comparisons with existing methods on various datasets and ablation studies under different constraints are provided to validate the effectiveness of our model.

\subsection{Implementation Details}
\mypara{Pre-training.}
 We pre-train the \ourmodel for 10 epochs on RVL-CDIP. RVL-CDIP is a dataset of real-world scanned documents that contains a small proportion of low-quality images. Current research in the visual document understanding field generally utilizes the raw OCR annotations provided in RVL-CDIP-EasyOcr. To align with other methods, we did not apply stricter filtering to the OCR results. We use AdamW optimizer with learning rate 2e-5, linear decay learning rate scheduler, batch size 32, and weight decay of 0.1. The maximum sequence length of the encoder and decoder is 512 and 128, respectively. We use pre-trained LayoutLMv3~\cite{huang2022layoutlmv3} to initialize the \encoder and keep it frozen during pre-training.
 Following~\cite{huang2022layoutlmv3}, the image resolution is set to 224$\times$224. Generally, the words within an input image should not be too long or too short, thus the images with less than 5 or more than 512 word-level tokens, are skipped. 
 For each sample in a batch, we randomly generate 8 instructions for each of the pre-training strategies mentioned in~\cref{subsec_pretrain_intro}.

\mypara{Fine-Tuning.}
We finetune the \ourmodel for 200 epochs on CORD and FUNSD, and 100 epochs on POIE. We train our model for 5000 steps under few-shot settings. We fine-tuned all methods (including LayoutLMv3 and GeoLayoutLM) on the DocILE dataset before conducting few-shot experiments. All parameters of our model are trainable during fine-tuning. Optimizer configuration is similar to pre-training but with a batch size of 4.

\subsection{Datasets}
\mypara{RVL-CDIP}~\cite{harley2015evaluation} includes 400k grayscale images from 16 categories, with 25k for each class. The train, validation, and test splits contain 320k, 40k, and 40k samples, respectively.

\mypara{DocILE}~\cite{vsimsa2023docile} comprises three subsets: an annotated set, an unlabeled set, and a synthetic set. Only the annotated set is used in this work, which contains 6,680 real business documents from publicly available sources.

\mypara{FUNSD}~\cite{jaume2019funsd} is a benchmark dataset specifically designed for Form understanding, and it has 149 and 50 samples for train and test. We evaluate the entity extraction task: predicting the entity values given the entity names such as "question", "answer", "header", or "other". The evaluation metric employed is field-level F1, which measures the accuracy of the extracted entities. The same evaluation metric is used for the following datasets as well.

\mypara{CORD}~\cite{park2019cord} stands for the Consolidated Receipt benchmark for post-OCR parsing. It has 1,000 receipt data samples with 30 fine-grain entities from 4 categories such as "total" or "subtotal".
The train, validation, and test splits contain 800, 100, and 100 samples, respectively. 

\mypara{POIE}~\cite{kuang2023visual} is a public benchmark for OCR and information extraction from real-scene products. Within the POIE dataset, there are 21 entity categories pertaining to nutritional information that need to be extracted. The dataset comprises 2,250 training images and 750 test images. 

\subsection{Main Results}

\begin{table*}[ht]
\centering
\setlength{\tabcolsep}{3.5mm}
\begin{adjustbox}{max width=1.\textwidth}
\begin{tabular}{lccccccccccc}
\toprule
\multirow{2}{*}{Method} & \multirowcell{1}{Generative \\ Method} & \multirow{2}{*}{Modality} & \multicolumn{3}{c}{FUNSD} & \multicolumn{3}{c}{CORD} & \multicolumn{3}{c}{POIE} \\
\cmidrule(lr){4-6} \cmidrule(lr){7-9} \cmidrule(lr){10-12}
&         &         & P     & R     & F     & P     & R     & F &P     & R     & F\\
\midrule
BERT\textsubscript{large}~\cite{huang2022layoutlmv3}   &    & T      & -     & -     & 65.63  & -     & -     & 90.25  & -     & -     & -\\
SPADE~\cite{hwang2021spatial}                              &      &  T+L        & -     & -     & 70.5  & -     & -     & 91.5  & -     & -     & -\\
GraphRevisedIE~\cite{cao2023graphrevisedie}     &      & T+L         & 76.67 & 80.22 & 78.41 & 93.91 & 94.61 & 94.26 & - & - & - \\
LayoutLMv2\textsubscript{base}~\cite{xu2021layoutlmv2}     &    & T+L+I          & 80.29 & 85.39 & 82.76 & 94.53 & 95.39 & 94.95 & - & - & 84.18 \\
LayoutLMv2\textsubscript{large}~\cite{xu2021layoutlmv2}    &      & T+L+I         & 83.24 & 85.19 & 84.20 & 95.65 & 96.37 & 96.01 & -     & -     & - \\
TILT\textsubscript{large}~\cite{powalski2021going}         & $\checkmark$ & T+L+I & -     & -     & -     & -     & -     & 96.33 & -     & -     & - \\ 
DocFormer~\cite{appalaraju2021docformer}                   &       & T+L+I       & 82.29 & 86.94 & 84.55 & 97.25 & 96.74 & 96.99 & -     & -     & -\\
UniDoc~\cite{gu2021unidoc}                                 &        & T+L+I      & -     & -     & 87.96 & -     & -     & 96.64 & -     & -     & -\\ 
StrucText~\cite{li2021structext}                           &      & T+L+I        & 85.68 & 80.97 & 83.09 & -     & -     & -     & -     & -     & -\\ 
BROS\textsubscript{base}~\cite{hong2022bros}               &     & T+L         & 81.16 & 85.02 & 83.05 & -     & -     & 96.50 & -     & -     & -\\ 
Gbada et al.~\cite{gbada2024multimodal}                      &     & T+L+I         & -     & -     & 80.40 & -     & -     & 97.20 &-         &-       & -    \\
BROS\textsubscript{large}~\cite{hong2022bros}              &      & T+L        & 82.81 & 86.31 & 84.52 & -     & -     & 97.28 & -     & -     & -\\ 
LiLT~\cite{wang2022lilt}                                   &      & T+L        & 87.21 & 89.65 & 88.41 & 95.98 & 96.16 & 96.07 & -     & -     & -\\ 
LayoutLMv3\textsubscript{base}~\cite{huang2022layoutlmv3}  &       & T+L+I        & 89.55 & 91.65 & 90.29 & -     & -     & 96.56 & -     & -     & -\\
LayoutLMv3\textsubscript{large}~\cite{huang2022layoutlmv3} &        & T+L+I       & 92.19 & 92.10 & 92.15 & -     & -     & 97.46 & 87.09 & 89.67& 88.36\\
GraphMLLM\textsubscript{large}~\cite{dai2024graphmllm} &        & T+L+I       &88.35& 88.70 & 88.52 & 96.20 & 96.56 & 96.38& -  & -  & - \\
ESP~\cite{yang2023modeling}                                &       & T+L+I        & -     & -     & 91.12 & -     & -     & 95.65 & -     & -     & -\\
GenKIE~\cite{cao2023genkie}                                & $\checkmark$  & T+L+I  & 83.45 & 83.45 & 83.45 & 95.75 & 95.75 & 95.75 & -     & -     & -\\
GeoLayoutLM~\cite{luo2023geolayoutlm}                      &       & T+L+I        & -     & -     & \ranksecond{92.86} & -     & -     & \rankfirst{97.97} & 90.25     & 92.14     & \ranksecond{91.19}\\
UDOP~\cite{tang2023unifying}                               & $\checkmark$ & T+L+I  & -     & -     & 91.62 & -     & -     & 97.58 & -     & -     & -\\ 
\midrule
\ourmodelshort(Ours)                                                      & $\checkmark$  & T+L+I  & 93.86 & 94.80 & \rankfirst{94.33} & 97.90 & 97.90 & \ranksecond{97.90} & 90.53 & 92.07 & \rankfirst{91.29}\\ 
\bottomrule
\end{tabular}
\end{adjustbox}
\caption{ \textbf{Comparisons with the SOTA in the full-shot setting}. 
“T/L/I” denotes “text/layout/image” modality. The highest and second highest results are marked with \textbf{bold} fonts and \underline{underlines}, respectively.
}
\label{tab:main_full_shot}
\end{table*}

\mypara{Comparisons with the SOTAs in the full-shot setting.}
We compare \ourmodel with previous OCR-dependent VIE methods in the full-shot setting, and the results are presented  in~\cref{tab:main_full_shot}. It can be observed that \ourmodel outperforms all existing methods on FUNSD, surpassing previous SOTA by
\textcolor{ForestGreen}{+1.47\%}. 
This proves that the matching mechanism can significantly benefit entity extraction, through the proposed dynamic weights generation. Compared to other generative methods, our approach has the advantage of only combining the recognized results of OCR, thus avoiding the instability in generating long texts. On CORD, our method achieves competitive performance, which is slightly lower than GeoLayoutLM by 0.07\%, but still ranks first among all generative methods. Given the annotation errors in the CORD dataset, we consider such differences to be negligible. We also conduct experiments on the scene text image dataset, POIE, and present the results in~\cref{tab:main_full_shot}. 
Notably, our model not only outperforms the baseline model LayoutLMv3 by \textcolor{ForestGreen}{+2.93\%} but also achieves slightly higher performance than GeoLayoutLM by 0.1\%. It demonstrates the superiority of our method in extracting information from natural scenes. 



\mypara{Evaluations in the Few-Shot Setting.}
We perform few-shot experiments with PLBR, LayoutLMv3 and GeoLayoutLM on CORD and POIE, and the results are shown in~\cref{tab:main_few_shot_cord_poie}. The base model of PLBR is BROS\textsubscript{large}, the results presented in the table comes from the original paper. In this experiment, we reimplement LayoutLMv3 and GeoLayoutLM using their official codes. Specifically, we conduct 1-shot, 2-shot, 5-shot, and 10-shot experiments with a five-fold random sampling with replacement. We also compare the two proposed matching modules, \ourmodelshort and \ourmodelseqshort, to study the advantages of sequential matching.  Overall, our method, whether using module \ourmodelshort or \ourmodelseqshort,  demonstrates a clear advantage in few-shot scenarios. Moreover, as the number of samples decreases, our method exhibits an increasingly greater advantage over GeoLayoutLM on CORD. 
We observe that our proposed model with sequential matcher obtains a decent margin over the other two methods on the 1-shot setting on CORD, resulting in an increase of \textcolor{ForestGreen}{+23.21\%} and \textcolor{ForestGreen}{+22.72\%} respectively. The practical value of 1-shot learning is huge, as a single image can serve as a template for quick deployment. Usually, we are able to obtain the rough structural topology and target keys from a single image at first glance. Under the 10-shot setting, the performance gap between our method and GeoLayoutLM is decreasing. However, compared to LayoutLMv3, both GeoLayoutLM and our method still maintain a 5\% advantage. This indicates that spatial perception pre-training tasks can be beneficial for few-shot VIE. In addition, the comparison of different matching modules in our method reveals that the sequential matcher consistently maintains an advantage of over 1\%. This indicates that sequential matching successfully matches more answers in the presence of limited instructions. The overall few-shot experiments demonstrate that the proposed \ourmodel, as compared to the classification-based method, is better able to leverage the prior knowledge of layout and prompts, enabling more effective learning from limited data.

\begin{table*}[h]
\small
\centering
\setlength{\tabcolsep}{1.3mm}
\begin{adjustbox}{max width=0.98\textwidth}
\begin{tabular}{l|cccc|cccc}
\toprule
\multirow{2}{*}{Method} & \multicolumn{4}{c|}{CORD} & \multicolumn{4}{c}{POIE} \\
\cmidrule{2-5} \cmidrule{6-9} 
    & 1-shot & 2-shots & 5-shots & 10-shots & 1-shot & 2-shots & 5-shots & 10-shots \\ \midrule
PLBR~\cite{Guo2024plbr} & - & - & 73.54 $\pm$ 0.32 & 83.70 $\pm$ 0.37 & - & - & - & - \\
LayoutLMv3 & 38.95 $\pm$ 2.04 & 63.80 $\pm$ 2.02 & 74.59 $\pm$ 1.50 & 79.09 $\pm$ 0.79 & 14.71 $\pm$ 2.34 & 16.14 $\pm$ 1.97 & 21.69 $\pm$ 1.16 & 36.62 $\pm$ 0.99 \\
GeoLayoutLM & 39.44 $\pm$ 6.43 & 66.02 $\pm$ 5.73 & 79.61 $\pm$ 2.04 & 84.12 $\pm$ 2.34 & 17.65 $\pm$ 1.99 & 27.87 $\pm$ 4.31 & 30.52 $\pm$ 1.72 & 42.77 $\pm$ 2.51 \\ \midrule
\ourmodelshort(Ours) & 58.08 $\pm$ 5.35 & 74.85 $\pm$ 2.98 & 79.93 $\pm$ 3.41 & 84.71 $\pm$ 2.65 & 18.62 $\pm$ 0.27 & 18.73 $\pm$ 2.75 & 24.23 $\pm$ 3.09 & 34.71 $\pm$ 2.23 \\
\ourmodelseqshort(Ours) & \textbf{62.16 $\pm$ 4.12} & \textbf{75.97 $\pm$ 3.81} & \textbf{83.33 $\pm$ 0.87} & \textbf{85.88 $\pm$ 1.68} & \textbf{26.74 $\pm$ 1.11} & \textbf{28.67 $\pm$ 3.47} & \textbf{36.50 $\pm$ 1.29} & \textbf{45.00 $\pm$ 2.38} \\ \bottomrule
\end{tabular}
\end{adjustbox}
\vspace{-2mm}
\caption{\textbf{Performance in various few-shot settings.} Mean accuracy and standard deviation ($\pm$std) are reported over 1-/2-/5-/10-shots settings. \ourmodelseqshort denotes the \ourmodel with \seqmatcher.}
\label{tab:main_few_shot_cord_poie}
\vspace{-4mm}
\end{table*}

\mypara{Effectiveness of the Matcher and Pre-training in Few-shot Learning.}
GeoLayoutLM and our \ourmodel both utilize the pre-trained weights of LayoutLM-V3 for their visual encoding. The primary distinctions between these models and \ourmodelshort are twofold:
1) GeoLayoutLM employs geometric spatial pre-training, while \ourmodel designs contextual spatial pre-training, 2) For decoding, \ourmodel employs a matching strategy, while the other methods use direct classification. ~\cref{tab:ablation_overall} shows that pre-training improves the base model by a margin of 1.4\%. ~\cref{tab:main_few_shot_cord_poie} demonstrates that under settings with fewer than 5 samples, \ourmodel exhibits improvements significantly greater than 5\%. This suggests that the matching mechanism plays a crucial role in few-shot learning, with contextual pre-training serving as a secondary contributor. As elaborated in the Motivation section, limited training samples lead to an increase in unseen categories. Classification models face challenges in accurately predicting unseen categories, while the matching model leveraging a metric learning mechanism, can identify the most probable instances for these categories.

\mypara{Evaluations with Shuffled OCR Input.}
As mentioned previously, it is important to note that the correct reading order cannot always be guaranteed in the majority of real-world applications.
To evaluate the impact of such ordering issues, we conduct experiments using shuffled OCR.  We divide the blocks into word-level segments and then randomly shuffle the order of them.  
\seqmatchershort demonstrates superior ability in handling disordered OCR on both datasets consistently.
\seqmatchershort significantly surpasses GeoLayoutLM by \textcolor{ForestGreen}{+1.01\%}, \textcolor{ForestGreen}{+1.9\%} and \textcolor{ForestGreen}{+2.68\%} on FUNSD, CORD and POIE respectively (see~\cref{tab:main_shuffle}.), which indicates the effectiveness of the proposed \seqmatchershort in real-world applications with shuffled text. Note that the GeoLayoutLM also has a special design for spatial positioning, which gives it some resistance to shuffled text. When compared to LayoutLMv3, our method demonstrates a larger improvement. It achieved a respective increase of \textcolor{ForestGreen}{1.99\%}, \textcolor{ForestGreen}{6.7\%}, and \textcolor{ForestGreen}{4.96\%} on the three datasets.
Besides, \ourmodelseqshort exhibits better performance than \ourmodel, indicating that the sequential matcher is more tolerant to OCR shuffling than the BIO matcher.
Moreover, qualitative results are demonstrated in~\cref{fig:vis_all_shuffle}. 
We observe that \seqmatchershort is able to correctly extract the entities even in challenging scenarios where reading order is hard to acquire due to document distortion or complex layouts.
\begin{table}[t]
\centering
\begin{adjustbox}{max width=1\textwidth}
\begin{tabular}{lccccccccc}
\toprule
\multirow{2}{*}{Method} & \multicolumn{3}{c}{FUNSD} & \multicolumn{3}{c}{CORD} & \multicolumn{3}{c}{POIE} 
\\ \cmidrule(lr){2-4} \cmidrule(lr){5-7} \cmidrule(lr){8-10}
            & P & R & F & P & R & F & P & R & F \\ \midrule
LayoutLMv3\textsubscript{large}   & 69.71 & 71.04 & 70.37 & 86.82 & 90.19 & 88.47 & 85.65 & 87.68 & 86.14 \\
GeoLayoutLM  & 72.32 & 70.40 & 71.35 & 92.63 & 93.91 & 93.26 & 87.66 & 89.19 & 88.42 \\ \midrule
\ourmodelshort(Ours)  & 72.30 & 70.89 & 71.58 & 88.60 & 90.79 & 89.69 & 90.92 & 90.59 & 90.76 \\
\ourmodelseqshort(Ours) & 71.35 & 73.39 & \textbf{72.36} & 95.14 & 95.21 & \textbf{95.17} & 89.65 & 92.61 & \textbf{91.10}\\ \bottomrule
\end{tabular}
\end{adjustbox}
\vspace{-1mm}
\caption{\textbf{Experimental results with shuffled OCR input.}}    
\label{tab:main_shuffle}
\vspace{-4mm}
\end{table}


\section{Analysis}
\subsection{Training and Inferring Efficiency}

\mypara{Pre-trained Data Size and Duration.} 
We conduct a comprehensive comparison of pre-trained data size and model size with several typical methods mentioned above, with the results delineated in ~\cref{tab:cmp_params}. Compared to other methods, our approach utilizes the least amount of pre-training data. Specifically, We utilize the RVL-CDIP ( 400K ) and DocILE ( 6.7K )  datasets to enhance the model's contextual capabilities, in contrast to most other datasets that utilize the extensive IIT-CDIP (11M) pre-training data. The visual encoder in our approach leverages the pre-trained weights from LayoutLMv3, which are frozen in pre-training. The proposed pre-training strategy is applied exclusively to the language part ( target generator ), making the whole pre-training stage highly efficient, which is completed in 63 hours with four A100 GPUs.
\begin{table}[H]
\centering
\begin{adjustbox}{max width=0.7\textwidth}
\begin{tabular}{lcccc}
\toprule
    Method & Pre-trained Docs & Params & FUNSD-F & CORD-F \\ \midrule
BROS\textsubscript{large}   & 11M & 340M & 84.52 & 97.28 \\
DocFormer  & 5M & 536M & 84.55 & 96.99  \\
ESP  & 0.9M & 50M & 91.12 & 95.65  \\
LayoutLMv3\textsubscript{base}  & 11M & 113M & 90.29 & 96.56  \\
LayoutLMv3\textsubscript{large}  & 11M & 368M & 92.08 & 97.46  \\
GeoLayoutLM& 11M & 399M & 92.86 & 97.97  \\
\midrule
GC\textsubscript{base}(Ours)  & 0.4M & 362M & 93.15 & 97.64 \\
GC\textsubscript{large}(Ours)  & 0.4M & 623M & 94.33 & 97.90 \\
\bottomrule
\end{tabular}
\end{adjustbox}
\caption{\textbf{Comparisons of Pre-trained Data Size and Model Size.} FUNSD-F and CORD-F denote F-measure on FUNSD and CORD respectively.}    
\label{tab:cmp_params}
\vspace{-4mm}
\end{table}

\begin{figure*}[t]
\centering
\captionsetup[subfloat]{font=scriptsize}
{\includegraphics[width=0.6\linewidth]{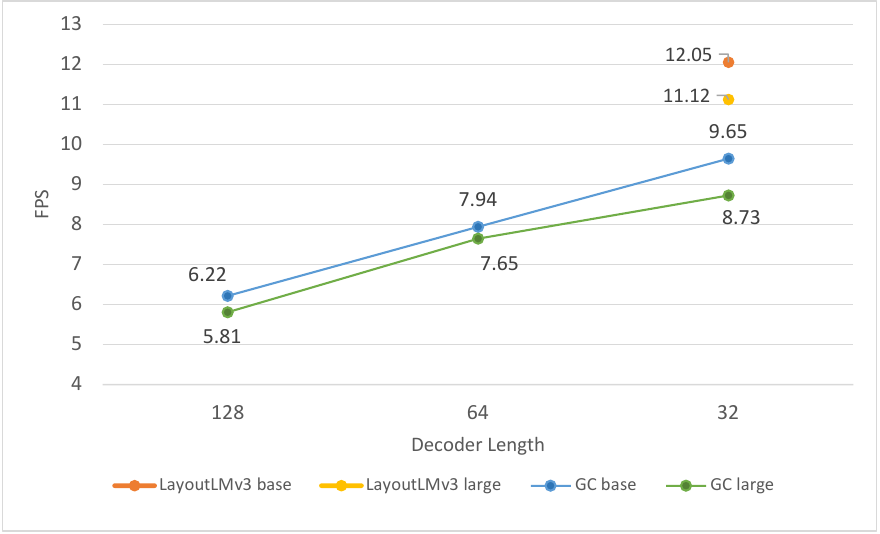}%
}
\vspace{-3mm}
\caption{\textbf{Illustrations of inference speed on CORD. The inference speed of \ourmodelshort is highly correlated with the length of the decoder. LayoutLMv3 does not have a decoder. Its efficiency is included here for comparison only.}}
\label{fig:vis_fps}
\vspace{-6mm}
\end{figure*}

\mypara{Model Size and Computation Cost.} We have developed a base version of \ourmodelshort that balances computational efficiency and performance. This model ranks second on FUNSD and third on CORD, as demonstrated in ~\cref{tab:cmp_params}. Notably, when compared to its larger counterpart, the base version achieves a 41.9\% reduction in model parameters, resulting in a total of 362 million parameters. \ourmodelshort primarily consists of three components: the visual encoder, resampler, and language decoder. By using the VIT-base in the base version, we reduce parameters in the visual part from 356M to 125M, significantly contributing to the overall reduction in the model's parameter count. 

In terms of inference speed, the proposed method demonstrates commendable performance. As shown in Fig.~\ref{fig:vis_fps}, GC\textsubscript{large} achieves 8.73 frames per second ( fps ), while GC\textsubscript{base} reaches 9.65 fps on CORD. Although these figures are slightly lower than those of LayoutLMv3, this discrepancy can primarily be ascribed to differences in model architecture. Specifically, LayoutLMv3 adopts an encoder followed by a classification layer, whereas \ourmodelshort employs an encoder-decoder-matcher mechanism. Given that the length of the decoding sequence, which corresponds to the categories of entities, directly influences the inference speed, we have evaluated the speed across various sequence lengths. The results indicate that a shorter sequence length enhances efficiency. In most cases, the number of entity categories is relatively limited, thus suggesting that a sequence length of 32 is theoretically applicable to the majority of scenarios. All experiments with different models are conducted on a single A100 GPU with a single batch and consistent versions of PyTorch and Transformers modules. It is noteworthy that the speed under the original LayoutLMv3 configuration is lower than that in Fig.~\ref{fig:vis_fps}, potentially due to the variations in PyTorch and Transformers versions.

\subsection{Ablation Studies}

\mypara{Effects of Pre-Training and Resampler.} 
We perform an ablation study on the two pivotal innovations of our method on FUNSD: pre-training strategies and \resampler, and the findings are presented in~\cref{tab:ablation_overall}. 
It can be observed that the introduction of pre-training strategies improves the performance of \unimatcher by \textcolor{ForestGreen}{+1.32\%}, while the resampler results in a boost of \textcolor{ForestGreen}{+0.73\%}. 
These results indicate a larger improvement in the pre-training strategies. 
Notably, the simultaneous utilization of the pre-training strategies and resampler yields the largest gain in performance.

\begin{table}[ht]
  \small
  \begin{minipage}{0.47\linewidth}
      \centering
    \begin{tabular}{c|cc|c}
    \toprule
    \# & PT & PAR & \ourmodelshort \\ \midrule
    0  &  &  & 92.74 \\
    1a & \checkmark &   & 94.06 \\
    1b & & \checkmark  & 93.47 \\ \midrule
    2  & \checkmark & \checkmark  & \textbf{94.33} \\ \bottomrule
    \end{tabular}
    \caption{\textbf{Ablations on pre-training and \resampler.} PT and PAR denote Pre-Training and \resampler.}
    \label{tab:ablation_overall}
  \end{minipage}
  \hfill
  \begin{minipage}{0.47\linewidth}
    \centering
    \vspace{3mm}
    \begin{tabular}{lc}
    \toprule
    Type & \ourmodelshort \\ \midrule
    Without Resampler & 92.92 \\
    Vanilla Resampler~\cite{alayrac2022flamingo} & 92.74 \\ \midrule
    PAR ( ours )        & \textbf{93.47} \\ \bottomrule
    \end{tabular}
    \vspace{3mm}
    \caption{\textbf{Ablations on different resampler designs on FUNSD.} PAR denotes \resampler.}
    \label{tab:ablation_resampler_design}       
  \end{minipage}

\end{table}

\mypara{Effects of different Resamplers.} We also compare different resampler designs for \ourmodel and the results are presented in~\cref{tab:ablation_resampler_design}. The vanilla resampler~\cite{alayrac2022flamingo} primarily focuses on visual feature distillation. We observe that \resampler outperforms the vanilla resampler by  \textcolor{ForestGreen}{+0.73\%} for \unimatcher. It confirms the conclusion that \resampler can promote matching by enhancing the input prompts and distilling the multi-modal embedding.

\begin{table}[th]
\centering
\small
\setlength{\tabcolsep}{3.5mm}
\setlength{\abovecaptionskip}{0.5em}
\begin{tabular}{c|ccc|cc}
\toprule
\# & MTF & SOD & SAD  & \ourmodelshort \\ \midrule
0  &     &     &     & 92.74 \\
1a & \checkmark & &  & 93.04 \\
1b & & \checkmark &  & 93.38 \\
1c & & & \checkmark  & 93.06 \\ \midrule
2a & \checkmark & \checkmark &  & 93.78 \\
2b & \checkmark &  & \checkmark  & 93.35 \\
2c & & \checkmark & \checkmark  & 93.41 \\ \midrule
3  & \checkmark & \checkmark & \checkmark & \textbf{94.06} \\ \bottomrule
\end{tabular}
\caption{\textbf{Ablations on pre-training strategies.} Experiments are conducted on the FUNSD dataset. }
\label{tab:ablation_pretrain}
\vspace{-5mm}
\end{table}

\mypara{Effects of Pre-Training Tasks.}
To investigate the effectiveness of three proposed pre-training tasks, we conduct ablative experiments on the FUNSD dataset without the prompt-aware resampler module and report the results in~\cref{tab:ablation_pretrain}.
In the first group, we find that pre-training either with MTF, SOD, or SAD task boosts the accuracy of the proposed GC. SOD has shown to be the most effective task, which indicates key-value pairs are more likely aligned either
horizontally or vertically.
In the second group, comparing 2a-2c with the baseline method, we observe that the combination of MTF and SOD is more effective with an improvement of \textcolor{ForestGreen}{+1.04\%}. By comparing the whole table, it can be observed that pre-training with all three tasks leads to the largest improvement, with the increases of \textcolor{ForestGreen}{+1.32\%}.

\begin{figure*}[th]
\centering
\captionsetup[subfloat]{font=scriptsize}
{\includegraphics[width=1\linewidth]{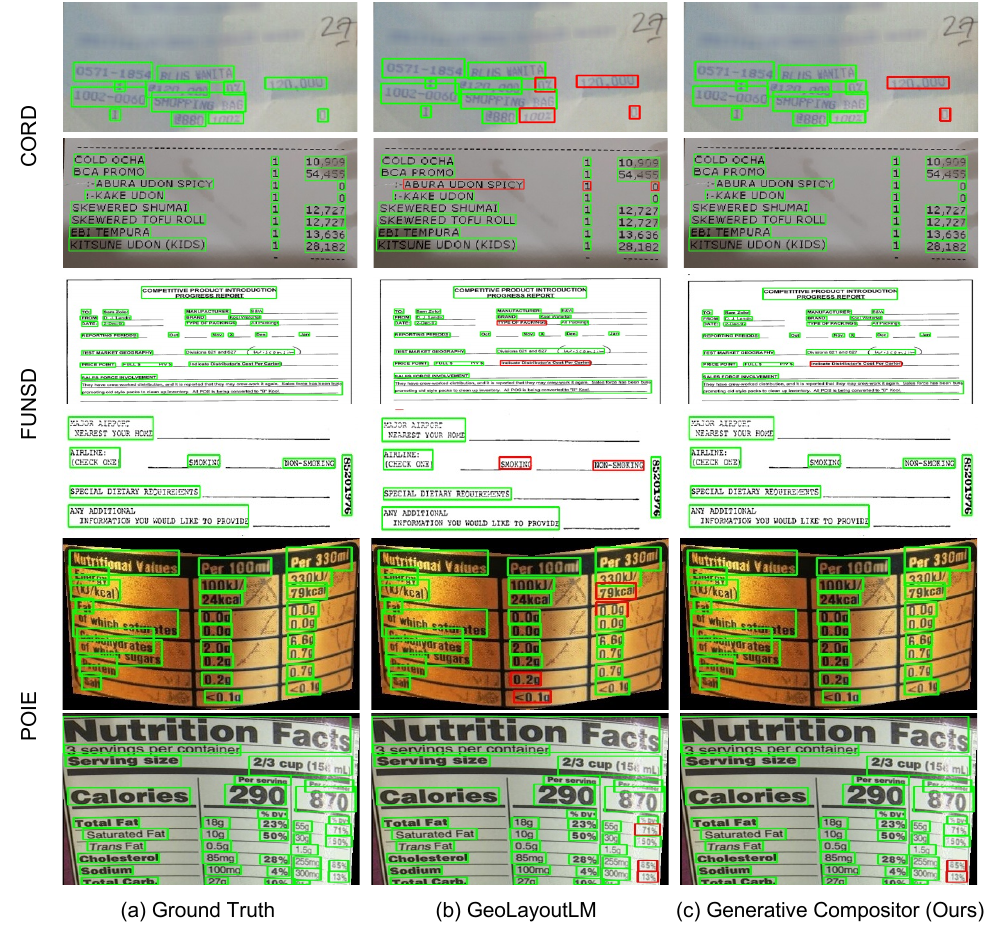}%
}
\caption{\textbf{Qualitative comparisons of GeoLayoutLM and \ourmodel on CORD, FUNSD and POIE}. The red box indicates incorrect VIE results.} 
\label{fig:vis_all}
\vspace{-3mm}
\end{figure*}

\subsection{Case Studies}
\mypara{Qualitative Analysis in full-shot Setting.}
Qualitative comparisons between GeoLayoutLM and our method on different datasets are presented in Fig.~\ref{fig:vis_all}. In the visualization results on CORD, it is evident that our method performs better in handling entities with shorter text. This may be attributed to the fact that our method learns visual information more effectively. However, similar to GeoLayoutLM, our \ourmodelshort fails when facing an instance that shows no explicit connection with other instances, for example, the first row of CORD.

In terms of real-world scenarios, as shown on POIE, VIE becomes more challenging for the layout information may be disturbed by nonlinear deformations. As shown in the first row of POIE, compared with GeoLayoutLM, our \ourmodelshort demonstrates a stronger ability against complex layouts and reading order errors.

\begin{figure*}[t]
\centering
\captionsetup[subfloat]{font=scriptsize}
{\includegraphics[width=0.95\linewidth]{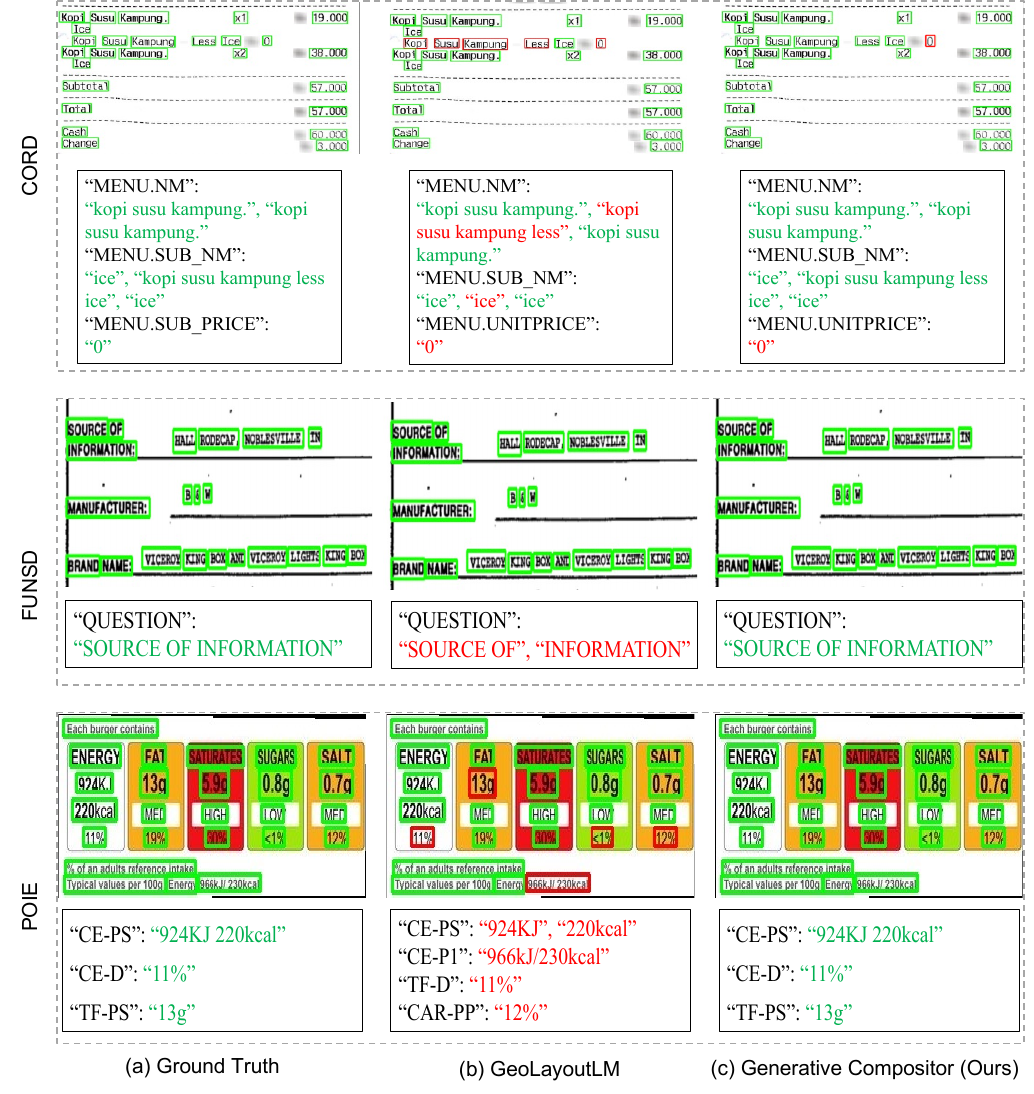}%
}
\vspace{-3mm}
\caption{\textbf{Qualitative results of GeoLayoutLM and \ourmodelseqshort with shuffled OCR}. The red boxes stand for incorrect classification of entities, while the red scripts stand for incorrect merge or split of entities.} 
\label{fig:vis_all_shuffle}
\vspace{-6mm}
\end{figure*}

\mypara{Qualitative Analysis with Shuffled OCR Input.} We divide the blocks into word-level segments and then
randomly shuffle the order of them. Thus, the scripts in the instance of CORD from Fig.~\ref{fig:vis_all_shuffle} may be changed from ``kopi susu kampung. X1 19.00'' to ``kopi subtotal 57.000. Cash Ice''. GeoLayoutLM is prone to have false positives of an entity or over-segment an entity, such as the case of ``kopi susu kampung less ice'' in the first row of Fig.~\ref{fig:vis_all_shuffle}, and ``SOURCE OF'' and `` INFORMATION'' in the second row of Fig.~\ref{fig:vis_all_shuffle} respectively.
Besides, when encountering complex structures, GeoLayoutLM is also prone to incorrectly link keys and values. For example, in the last row of Fig.~\ref{fig:vis_all_shuffle}, the linking of ``TF-D: 11\%'' and ``CAR-PP: 12\%'' are both incorrect links. Regarding our method, it is evident that it is able to against shuffled OCR, thanks to the implementation of the sequential matching module.


\section{Conclusions and Future Works}
In this paper we have introduced a novel generative-based method for Visual Information Extraction. The method adopts the idea of simultaneous retrieval and generation, which offers a new perspective for VIE. To enhance the few-shot capability of our proposed method, we have proposed three innovations: 1) Pre-training tasks to enhance the model’s perceptual capacity for spatial context information, 2) Prompt-aware resampler to distill the multi-modal embedding, 3) Sequential matcher to better leverage semantic information. Benefiting from the novel architecture and innovative modules, our model achieves state-of-the-art performance on VIE, particularly demonstrating significant improvement in the few-shot scenario. Besides, our model is also capable of handling input errors in OCR, such as disordered texts. Note that, the innovative modules are method-agnostic so that they can be applied to other frameworks.

Moving forward, there are two potential directions for future expansion: developing a multilingual variant of the proposed method and exploring its zero-shot capability. Compared to Latin, Chinese and Korean have wider character types and richer arrangements, which pose greater challenges for visual modules. Zero-shot VIE is widespread in applications, requiring the introduction of powerful language models. Both directions possess high scientific values, and we will leave them for future work.

\section{Acknowledgement}
This work was supported by the National Natural Science Foundation of China (No.62206013), and Alibaba Innovative Research (AIR) program.


\section*{}

\end{sloppypar}
\end{document}